\documentclass[11pt,a4paper]{article}
\usepackage[hyperref]{emnlp2018}
\usepackage{times}
\usepackage{latexsym}
\usepackage{graphicx}
\usepackage{url}
\usepackage{dirtytalk}
\usepackage{amsmath}
\usepackage{amssymb}
\usepackage{colortbl} 
\usepackage{booktabs}
\usepackage{caption}
\usepackage{array,multirow}

\aclfinalcopy 

\title{Angular-Based Word Meta-Embedding Learning}

\author{James O' Neill and Danushka Bollegala\\
  Department of Computer Science, University of Liverpool\\
  Liverpool, L69 3BX \\
  England \\
  {\tt \{james.o-neill,danushka.bollegala\}@liverpool.ac.uk} \\}

\date{13/08/2018}

\begin{document}
\maketitle

\begin{abstract}
Ensembling word embeddings to improve distributed word representations has shown good success for natural language processing tasks in recent years. These approaches either carry out straightforward mathematical operations over a set of vectors or use unsupervised learning to find a lower-dimensional representation. This work compares meta-embeddings trained for different losses, namely loss functions that account for angular distance between the reconstructed embedding and the target and those that account normalized distances based on the vector length. We argue that meta-embeddings are better to treat the ensemble set equally in unsupervised learning as the respective quality of each embedding is unknown for upstream tasks prior to meta-embedding. We show that normalization methods that account for this such as cosine and KL-divergence objectives outperform meta-embedding trained on standard $\ell_1$ and $\ell_2$ loss on \textit{defacto} word similarity and relatedness datasets and find it outperforms existing meta-learning strategies. 

\end{abstract}

\section{Introduction}

Meta-embeddings are a quick and useful prior step for improving word representations in natural language learning tasks. This involves combining several learned embeddings in a way that improve the overall input representation. This approach is a less computationally expensive compared to if a practitioner were to train a set of word embeddings from scratch, particularly when considering non-sliding window methods.
The most straightforward approaches to meta-embeddings are: concatenation (CONC) and averaging (AV). The former is limited since the dimensionality grows large for multiple embeddings as more vectors are concatenated and the latter, while fast, does not preserve most of the information encoded in each embedding when taking the arithmetic mean. Although, it would seem surprising concatenating vectors from different embedding spaces is valid, it has been shown that ~\cite{coates2018frustratingly} AV approximates CONC even though the embedding spaces are different. Although, to address the loss of information when using AV, Singular Value Decomposition has been used as a dimensionality reduction technique to factorize the embeddings into a lower-rank approximation of the concatenated meta-embedding set. 

Linear methods include the use of a projection layer for meta-embedding (known as 1TON) ~\newcite{yin2015learning}, which is simply trained using an $\ell_2$-based loss. Similarly, Bollegala et al. ~\cite{bollegala2017learning} has focused on finding a linear transformation between count-based and prediction-based embeddings, showing that linearly transformed count-based embeddings can be used for predictions in the localized neighborhoods in the target space. 

Most recent work ~\cite{bao2018meta} has focused on the use of an autoencoder (AE) to encode a set of $N$ pretrained embeddings using 3 different variants: (1) Decoupled Autoencoded Meta Embeddings (DAEME) that keep activations separated for each respective embedding input during encoding and uses a reconstruction loss for both predicted embeddings while minimizing the loss for each respective decoded output, (2) Coupled Autoencoded Meta Embeddings (CAEME) which instead learn to predict from a shared encoding and (3) Averaged Autoencoded Meta-Embedding (AAME) is simply an averaging of the embedding set as input instead of using a concatenation. This is the most relevant work to our paper, hence, we include these 3 autoencoding schemes along with aforementioned methods for experiments, described in Section \ref{sec:experiments}. We also include two subtle variations of the aforementioned AEs. The first predicts a target embedding from an embedding set using the remaining embedding set, post-learning the single hidden layer is used as the word meta-embedding. The second method is similar except an AE is used for each input embedding to predict the designated target embedding, followed by an averaging over the resulting hidden layers. This alternative is described in more detail in Section \ref{sec:method}. 

The aforementioned previously proposed unsupervised learning have a common limitation, that is minimising the Euclidean ($\ell_2$) distance between source word embeddings and their meta-embedding. \newcite{Arora:TACL:2016} have shown that the $\ell_2$ norm of a word embedding vector is proportional to the frequency of the corresponding word in the corpus used to learn the word embeddings. Considering that in meta-embedding learning we use source embeddings trained on different resources, we argue that it is more important to preserve the semantic orientation of words, which is captured by the angle between two word embeddings, not their length. Indeed, cosine similarity, a popularly used measure for computing the semantic relatedness among words, ignores the length related information. Additionally, we note the relationship between KL-divergence and cosine similarity in the sense both methods perform a normalization that is proportional to the semantic information. Hence, we compare several popular measures such as MSE and MAE that use $\ell_2$ and $\ell_1$ respectively, against KL-divergence and cosine similarity for the purpose of learning meta-embeddings and show that the loss which accounts for this orientation consistently outperforms the former objectives that only consider length. We demonstrated this across multiple benchmark datasets.

\section{Methodology}\label{sec:method}
\begin{figure}[ht]
\begin{center}
 \includegraphics[scale=0.75]{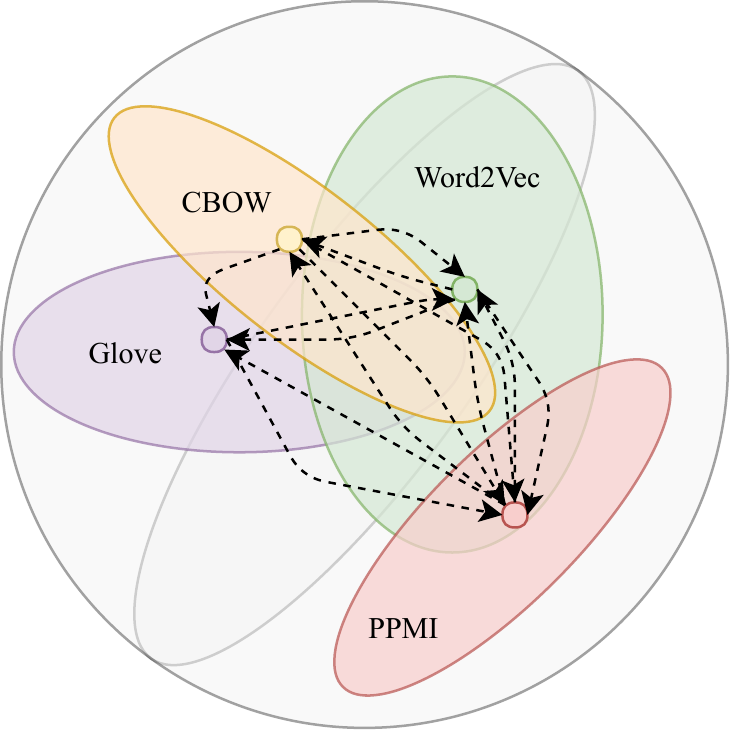}
 \caption{TAE Embedding}\label{fig:mre}
\end{center}
\end{figure}

Before describing the loss functions used, we explain the aforementioned variation on the autoencoding method and how it slightly differs from 1TON/1TON$^{+}$ ~\cite{yin2015learning} and standard AEs ~\cite{bao2018meta} presented in previous work. Target Autoencoders (TAE) are defined as learning an ensemble of nonlinear transformations between sets of bases $X_{s}$ in sets of vector spaces $\mathcal{X}_{S} = \{\mathcal{X}_{1},..,\mathcal{X}_{s},..,\mathcal{X}_{N}\} \quad s.t \quad \mathcal{X}_{s}\in\mathbb{R}^{|v_{s}|\times d_{s}} $ to a target space $\mathcal{X}_{t} \in \mathcal{R}^{|v_{t}|\times d_{t}}$, where $\mathtt{f}_{w}^{(i)}: \mathcal{X}_{S}^{(i)}$ $\to$ $\mathcal{X}_{t} \quad \forall i $ is the nonlinear transformation function used to make the mapping. Once a set of $M$ number of parameteric models $\mathtt{f}_{w} = \{\mathtt{f}_{w}^{(1)}, \mathtt{f}_{w}^{(i)}, ..,\mathtt{f}_{w}^{(M)}\}$ are trained with various objective functions to learn the mappings between the word vectors we obtain a set of lower-dimensional target latent representation that represents different combinations of mappings from one vector space to another. After training, all $H$ set of latent variables $Z_{s}=\{z_{1},..,z_{H}\}$ are concatenated with an autoencoded target vector. This means thats all vector spaces have been mapped to a target space and there hidden meta-word representations have been averaged, as illustrated in Figure \ref{fig:mre}. 

\begin{figure}[!bp]
\begin{center}
 \includegraphics[scale=0.45]{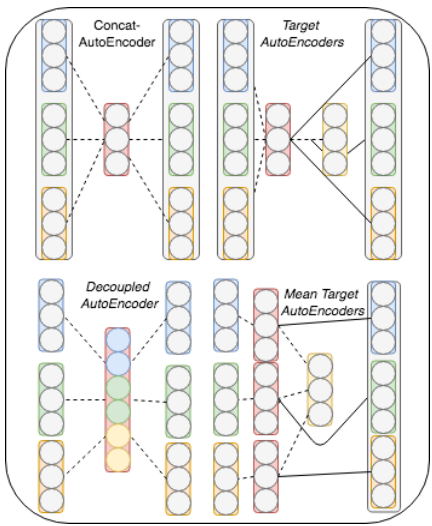}
 \caption{AE Meta-Embedding Methods}\label{fig:meta_methods}
\end{center}
\end{figure}

Figure \ref{fig:meta_methods} shows a comparison of the previous autoencoder approaches ~\cite{bao2018meta} (left) and the alternative AE (right), where dashed lines indicate connections during training and bold lines indicate prediction. The Concat-AutoEncoder (CAEME) simply concatenates the embedding set into a single vector and trains the autoencoder so to produce a lower-dimensional representation (shown in red), while the decoupled autoencoder (DAEME) keeps the embedding vectors separate in the encoding. In contrast the target encoder (TAE) is similar to that of CAEME only the label is a single embedding from the embedding set and the input are remaining embeddings from the set. After training, TAE then concatenates the hidden layer encoding with the original target vector. The Mean Target AutoEncoder (MTE) instead performs an averaging over separate autoencoded representation. 

\paragraph{AutoEncoder Settings}
The standard Autoencoder (AE) is a 1-hidden layer AE of hidden layer dimension $h_{d}=200$. Weights are initialized with a normal distribution, mean $\mu=0$ and standard deviation $\sigma= 1$. Dropout is used with a dropout rate $p=0.2$ for all datasets. The model takes all unique vocabulary terms pertaining to all tested word association and word similarity datasets ($n = 4819$) and performs Stochastic Gradient Descent (SGD) with batch size $\tilde{x} = 32$ trained between 50 epochs for each dataset $\forall d \in \mathcal{D}$. This results in a set of vectors $X_{j} \in \mathbb{R}^{|v_{j}| \times 200} \; \forall j$ that are then used for finding the similarity between word pairs. The above parameters were chosen ($h_d$, $\tilde{x}$ and number of epochs) over a small grid search. As stated, we compare against previous methods ~\cite{yin2015learning,bao2018meta} that use $\ell_2$ distance, as shown in Equation \ref{eq:mse_loss}).

\begin{equation}\label{eq:mse_loss}
	\mathcal{L}_{\theta}(\hat{y},y) = \frac{1}{N}\sum_{i=1}^{N} (y^{(i)} - \hat{y}^{(i)})^{2}
\end{equation}

Similarly, the Mean Absolute Error ($\ell_1$ norm of difference) loss $\frac{1}{N}\sum_{i=1}^{N}|y - \hat{y}|$ is tested. We also compare against a KL divergence objective, as shown in Equation \ref{eq:kl_divergence}, $\hat{y}$ is the last activation output from the log-softmax that represents $q(x)$ and the KL-divergence is given as $KL(p|q) = \sum_{i=1}^{N}p(x_i)\log \big( q(x_i)/p(x_i)\big)$. 

\begin{equation}\label{eq:kl_divergence}
   \mathcal{L}_{\theta}(\hat{y},y) = \frac{1}{N}\sum_{i=1}^{N} y^{(i)} \cdot \left( \log \big(y^{(i)}\big) - \hat{y}^{(i)} \right)
\end{equation}

Since $\tanh$ functions are used and input vectors are $\ell_2$ normalized we propose a Squared Cosine Proximity (SCP) loss, shown in Equation \ref{eq:cosine_loss} where $m$ is the output dimensionality. This forces the optimization to tune weights such that the rotational difference between the embedding spaces is minimized, thus preserving semantic information in the reconstruction. In the context of its utility for the TAE, we too want to minimize the angular difference between corresponding vectors in different vector spaces. It is also a suitable fit since it is a proper distance measure (i.e symmetric), unlike KL-divergence. 

\begin{equation}\label{eq:cosine_loss}
	\mathcal{L}_{\theta}(\hat{y},y) = \frac{1}{N}\sum_{i=1}^{N}\Bigg(1 - \frac{\sum_{j=1}^{m}\hat{y}_{ij} * y_{ij}}{\sqrt{\sum_{ij}^{m} \hat{y}_{ij}^{2}}\sqrt{\sum_{ij}^{m}y_{ij}^{2}}}\Bigg)^{2}
\end{equation}

The model is kept relatively simple so that the comparisons against previous methods are directly comparable and that the performance comparison between the proposed SCP loss and KL divergence loss against MSE and MAE is controlled. Additionally, all comparison are that of models which are only trained from co-occurrence statistics that are not leveraging any external knowledge (e.g AutoExtend ~\cite{rothe2015autoextend}). 

\section{Experiments}\label{sec:experiments}
The following word association and word similarity datasets are used throughout experimentation: Simlex ~\cite{hill2015simlex}, WordSim-353 \cite{finkelstein2001placing}, RG ~\cite{rubenstein1965contextual}, MTurk (MechanicalTurk-771)~\cite{halawi2012large}, RareWord (RW) ~\cite{luong2014addressing} and MEN ~\cite{bruni2012distributional}. The word vectors considered in the embeddings set are $\mathtt{skipgram}$ and $\mathtt{cbow}$ ~\cite{mikolov2013efficient}, FastText ~\cite{bojanowski2016enriching}, LexVec ~\cite{salle2016matrix}, Hellinger PCA (HPCA)~\cite{lebret2013word} and Hierarchical Document Context (HDC) ~\cite{sun2015learning}. We now report results on the performance of meta-embedding autoencodings with various loss functions, while also presenting target autoencoders for combinations of word embeddings and compare against existing current SoTA meta-embeddings.

Table \ref{tab:meta_results} shows the scaled Spearman correlation test scores, where (1) shows the original single embeddings, (2) results for standard meta-embedding approaches that either apply a single mathematical operation or employ a linear projection as an encoding, (3) presents the results using autoencoder schemes by ~\cite{bao2018meta} that we have used to test the various losses, (4) introduces TAE without concatenating the target $Y$ embedding post-training with MSE loss and (5) shows the results of concatenating $Y$ with the lower-dimensional (200-dimensions) vector that encodes all embeddings apart from the target vector. Therefore reported results from (4) are of a 200d vector, while (5) concatenates the vector leading to a vector between 300-500 dimensions dependent on the target vector. All trained encodings from sections 3-5 are 200-dimensional vectors. Results in red shading indicate the best performing meta-embedding for all presented approaches, while black shading indicates the best performing meta-embedding for the respective section. 

Best performing word meta-embeddings are held between concatenated autoencoders that use the proposed Cosine-Embedding loss, while a KL-divergence also performs well on Simlex and RareWord. Interestingly, both of these dataset are distinct in that Simlex is the only dataset providing scores on \textit{true similarity} instead of free association, which has shown to be more difficult for word embeddings to account for ~\cite{hill2016simlex}, while Rareword provides morphologically complex words to find similarity between. Concretely, it would seem KL-divergence is well suited for encoding when the word relations exhibits of a more complex or rare nature. Similarly, we find SCP loss to achieve best results on RG and MEN, both the smallest and largest datasets of the set.
Furthermore, the TAE variant has lead to competitive performance overall against other meta-embedding approaches and even produces best results on WS353. Lastly, we find that HPCA embeddings are relatively weak for word similarity.

\begin{table}[ht]
\centering
\captionsetup{justification=centering, margin=0cm}

\resizebox{0.95\linewidth}{!}{%
\begin{tabular}{c|ccccccc}
 \toprule
\multicolumn{1}{l}{1. Embeddings} & Simlex & WS353 & RG & MTurk & RW & MEN \\
\midrule

Skipgram & \cellcolor{black!20}44.19 & \cellcolor{black!20}77.17 & 76.08 & 68.15 & \cellcolor{black!20}49.70 & 75.85\\
FastText & 38.03 & 75.33 & 79.98 & 67.93 & 47.90 & 76.36 \\
GloVe & 37.05 & 66.24 & 76.95 & 63.32 & 36.69 & 73.75 \\
LexVec & 41.93 & 64.79 & 76.45 &  \cellcolor{black!20}71.15 & 48.94 & \cellcolor{black!20}80.92\\
HPCA & 16.60 & 57.11 & 41.72 & 37.45 & 13.36 & 34.90 \\
HDC & 40.68 & 76.81 & \cellcolor{black!20}80.58 & 65.76 & 46.34 & 76.03\\

\midrule
\multicolumn{1}{l}{2. Standard Meta} & & & & & & & \\
\midrule

CONC & \cellcolor{black!20}42.57 & \cellcolor{black!20}72.13 & \cellcolor{black!20}81.36 & \cellcolor{black!20}71.88 & 49.91 & 80.33 \\
SVD & 41.10 & 72.06 & 81.18 & 71.50 & 49.13 & 79.85 \\
AV & 40.63 & 70.5 & 80.05 & 70.51 & 49.28 & 78.31 \\
1TON & 41.30 & 70.19 & 80.20 & 71.52 & 50.80 & \cellcolor{black!20}80.39 \\
1TON* & 41.49 & 70.60 & 78.40 & 71.44 & \cellcolor{black!20}50.86 & 80.18\\

\midrule
\multicolumn{1}{l}{3. $\ell_2$-AE} & & & & & & & \\
\midrule

Decoupled & 42.56 & 70.62 & 82.81 & 71.16 & 50.79 & 80.33 \\
Concatenated & \cellcolor{black!20}43.10 & \cellcolor{black!20}71.69 & 84.52 & \cellcolor{red!20}71.88 & \cellcolor{black!20}50.78 & 81.18 \\

\midrule
\multicolumn{1}{l}{$\ell_1$-AE} & & & & & & & \\
\midrule

Decoupled & 43.52 & 70.30 & \cellcolor{black!20}82.91 & \cellcolor{black!20}71.43 & 51.48 & \cellcolor{black!20}81.16 \\
Concatenated & \cellcolor{black!20}44.41 & \cellcolor{black!20}70.96 & 81.16 & 69.63 & \cellcolor{black!20}51.89 & 80.92 \\

\midrule
\multicolumn{1}{l}{Cosine-AE} & & & & & & & \\
\midrule

Decoupled & 43.13 & 71.96 & 84.23 & 70.88 & 50.20 & 81.02 \\
Concatenated & \cellcolor{black!20}44.85 & \cellcolor{black!20}72.44 & \cellcolor{red!20}85.41 & 70.63 & \cellcolor{black!20}50.74 & \cellcolor{red!20}81.94 \\

\midrule
\multicolumn{1}{l}{KL-AE} & & & & & & & \\
\midrule
Decoupled & 44.13 & 71.96 & 84.23 & \cellcolor{black!20}70.88 & 50.20 & 81.02 \\
Concatenated & \cellcolor{red!20}45.10 & \cellcolor{black!20}74.02 & \cellcolor{black!20}85.34 & 67.75 & \cellcolor{red!20}53.02 & \cellcolor{black!20}81.14 \\

\midrule
\multicolumn{1}{l}{4. TAE} & & & & & & & \\
\midrule

$\to$Skipram & 37.80 & 67.33 & 76.50 & 63.41 & 37.52 & 74.86 \\
$\to$ FastText & 38.17 & 66.62 & 77.184 & 64.73 & 37.84 & 74.77 \\
$\to$Glove & 39.95 &  \cellcolor{red!20}77.14 & \cellcolor{black!20}81.58 & \cellcolor{black!20}68.82 & \cellcolor{black!20}47.94 & \cellcolor{black!20} 76.67 \\
$\to$ LexVec & 37.48 & 67.19 & 75.98 & 63.96 & 37.70 & 74.75 \\
$\to$HPCA & \cellcolor{black!20}40.78 & 65.79 & 38.64 & 59.49 & 38.65 & 74.50 \\
$\to$HDC & 38.15 & 66.96 & 76.62 & 63.08 & 37.53 & 76.62  \\

\midrule
\multicolumn{1}{l}{5. TAE +$Y$} & & & & & & & \\
\midrule


$\to$ Skipram & 42.43 & 75.33 & 80.11 & 66.51 & 44.77 & 78.98 \\
$\to$ FastText & 41.69 & 72.65 & 80.51 & 67.64 & 47.41 & 77.48  \\
$\to$ Glove & 41.75 & \cellcolor{black!20}76.65 & \cellcolor{black!20}82.40 & 68.92 & \cellcolor{black!20}48.83 & 78.27 \\
$\to$ LexVec & \cellcolor{black!20}42.85 & 73.33 & 80.97 & \cellcolor{black!20}69.17 & 46.71 & \cellcolor{black!20}79.63 \\
$\to$ HPCA & 40.03 & 69.65 & 70.43 & 61.31 & 36.38 & 73.10 \\
$\to$ HDC & 42.43 & 74.08 & 80.11 & 66.51 & 44.76 & 77.93  \\

\bottomrule
\end{tabular}%
}

\vspace{1em}
   \caption{Meta-Embedding Results}
  \label{tab:meta_results}
\end{table}

\section{Conclusion}
We find the meta-embeddings trained using Autoencoders with a Squared Cosine loss and a KL-divergence loss improves performance in the majority of cases, reinforcing the argument that accounting for angles explicitly through normalization (log-softmax for KL) is an important criterion for encoding. It is particularly useful for distributed word representations, since embeddings are learned from large documents of varying length and semantics. Lastly, we have shown its use in the context of improving meta-embeddings, although this suggests cosine loss is also suitable for minimizing angular differences for word embeddings, not only for meta-embeddings. Concretely, this paper has carried out a comprehensive study of methods to embed a lower-dimensional representation from embedding sets, while proposing losses that explicitly keep angular information intact for meta-embeddings.

\bibliography{emnlp2018}
\bibliographystyle{acl_natbib_nourl}

\appendix

\end{document}